\documentclass{Interspeech}
\usepackage{tabularx}
\usepackage{booktabs}

\usepackage{ragged2e}
\usepackage[utf8]{inputenc} 
\usepackage[T5]{fontenc} 
\usepackage{xcolor}
\usepackage{hyperref}

\interspeechcameraready

\title{Dynamic Context-Aware Streaming Pretrained Language Model For Inverse Text Normalization
}

\author[affiliation={1,2,3,*}]{Luong}{Ho}
\author[affiliation={1,*}]{Khanh}{Le}
\author[affiliation={1}]{Vinh}{Pham}
\author[affiliation={1}]{Bao}{Nguyen}
\author[affiliation={1}]{Tan}{Tran}
\author[affiliation={2,3}]{Duc}{Chau}

\affiliation{}{Zalo AI}{Vietnam}
\affiliation{}{University of Science}{Ho Chi Minh City, Vietnam}
\affiliation{}{Vietnam National University}{Ho Chi Minh City, Vietnam}
\email{\{congluong2122001,khanhld218,phamvinh257,baonguyenduy2002,tranctan96\}@gmail.com, ctduc@fit.hcmus.edu.vn}


\usepackage{comment}
\usepackage{graphicx}
\usepackage{multirow}
\usepackage{textcomp}
\usepackage{xcolor}
\usepackage{algorithm}
\usepackage{algpseudocode}

\begin{document}

\maketitle

\begin{abstract}
Inverse Text Normalization (ITN) is crucial for converting spoken Automatic Speech Recognition (ASR) outputs into well-formatted written text, enhancing both readability and usability.
Despite its importance, the integration of streaming ITN within streaming ASR remains largely unexplored due to challenges in accuracy, efficiency, and adaptability, particularly in low-resource and limited-context scenarios.
In this paper, we introduce a streaming pretrained language model for ITN, leveraging pretrained linguistic representations for improved robustness. To address streaming constraints, we propose Dynamic Context-Aware during training and inference, enabling adaptive chunk size adjustments and the integration of right-context information. Experimental results demonstrate that our method achieves accuracy comparable to non-streaming ITN and surpasses existing streaming ITN models on a Vietnamese dataset, all while maintaining low latency, ensuring seamless integration into ASR systems.
\end{abstract}

\section{Introduction}
Inverse Text Normalization (ITN) is a natural language processing task that converts spoken-language transcriptions, typically produced by automatic speech recognition (ASR) systems, into their proper written form. For example, the spoken phrase ``three point five dollars'' may need to be transformed into ``\$3.50''. ITN is essential for improving text accuracy and readability in applications like document generation, voice assistants, and machine translation. The task involves resolving linguistic variations in numbers, dates, currencies, and more, often requiring contextual understanding to handle ambiguities.

Recent advancements in ITN methods emphasize end-to-end neural network-based models, typically employing sequence-to-sequence (seq2seq) architectures \cite{10094599, saito-etal-2017-improving, choi24_interspeech}, which directly convert spoken-form input into written-form output. However, standard seq2seq models that rely on content-based attention often face challenges, including insertion, substitution, and deletion errors \cite{sunkara2021neural, antonova22_interspeech}. The most significant drawback of these systems is their tendency to produce unrecoverable errors that, while linguistically coherent, fail to preserve the original information accurately. The hybrid approach addresses these issues by combining a neural tagging model with finite state transducer (FST) based rules \cite{10094599, antonova22_interspeech, tan2023four, pusateri17_interspeech}, leveraging the adaptability of neural models and the precision of FST. This method enhances accuracy by mitigating common errors, preserving unmodified text, and efficiently handling overlapping source and target sentences. Its modular design improves scalability, allowing for easy rule adjustments without retraining the neural model, making it highly effective for ITN tasks. However, the hybrid approach requires significant effort to label inside–outside–beginning (IOB) tags for each word in a sentence, demanding considerable time and human resources. Although seq2seq methods also face data challenges, training them in an end-to-end manner by converting target sequences to source sequences generally involves less effort compared to the intensive tagging required for hybrid methods. To address the scarcity of training data, particularly in-domain data, recent advanced approaches have begun leveraging pre-trained language models \cite{choi24_interspeech, sunkara2021neural, antonova22_interspeech, tan2023four, singhal-etal-2023-scaling} or adopting data-driven strategies \cite{saito-etal-2017-improving, pusateri17_interspeech, paul22_interspeech}.

Integrating ITN into a streaming ASR system can be approached in two ways. The first is applying ITN after a silence period or end of speech, which has been the focus of most previous research. The second, more challenging approach, involves applying ITN in real-time as the user speaks, a scenario that has received limited attention in publications. In streaming ITN, seq2seq models face disadvantages due to their inability to operate in a streaming manner. Their encoder-decoder architecture, reliant on cross-attention, makes real-time processing inherently difficult. To date, only \cite{gaur2023streaming} has introduced a streaming ITN method to the community. However, their approach faces significant challenges, particularly in scenarios with limited data, such as conversational data, as their experiments are conducted in a large dataset setting. Additionally, their method relies on a fixed chunk size for processing, which does not align with variable ASR outputs. In practice, ASR systems may generate more or fewer words than the predefined chunk size of a streaming ITN model, resulting in cumulative unprocessed words that negatively impact the real-time experience. This fixed chunk size limitation is particularly problematic for tasks like punctuation restoration. For instance, when processing the phrase ``are you ok peter'' with a chunk size of three, the output may incorrectly become ``Are you OK? peter'' instead of the intended ``Are you OK, Peter?''.

In this paper, we propose a novel approach for streaming ITN tasks that achieves results comparable to non-streaming methods, even with limited datasets, while also addressing the challenge of insufficient surrounding context in chunk-based processing of complex sentences. 
Specifically:

1. Streaming Pretrained Language Model: We are the first to introduce a streaming ITN approach that leverages a pretrained language model. By modifying the architecture of the pretrained model, we enable real-time processing while taking advantage of the pretrained weights.

2. Dynamic Context-Awared Training and Inference: We propose a novel method to enhance the model's robustness across varying context sizes during both training and inference. This approach allows the model to process variable chunk sizes and right context lengths, unlike the fixed chunk size used in \cite{le24_interspeech}. This flexibility enables instant inverse normalization during speech, thus improving performance and user experience.

\section{Streaming Inverse Text Normalization With Pretrained Language Models}
\subsection{Pre-trained Language Models}
Bidirectional Encoder Representations from Transformers (BERT) \cite{devlin-etal-2019-bert} leverages the Transformer architecture to learn deep and bidirectional representations across all layers. During pre-training, BERT is trained on a massive, unlabeled text corpus using two unsupervised objectives: Masked Language Modeling and Next Sentence Prediction. These tasks enable the model to learn rich contextual representations. Subsequently, the pre-trained BERT model is fine-tuned on task-specific labeled datasets. This involves initializing the model with the pre-trained parameters and updating them to optimize performance on downstream tasks such as Question Answering and Natural Language Inference. 

In the context of Vietnamese language modeling, PhoBERT \cite{nguyen-tuan-nguyen-2020-phobert}, a BERT-based model trained on a 20GB word-level Vietnamese corpus, has achieved state-of-the-art performance. It excels in diverse Vietnamese NLP tasks, including word-level tasks like Part-of-Speech tagging, Dependency parsing, and Named-Entity Recognition, as well as the more complex Natural Language Inference task. PhoBERT, with 135 million parameters, leverages 12 encoder layers, 12 attention heads, and a 768 hidden size, crucially incorporating a Byte-Pair Encoding tokenizer optimized for the nuances of the Vietnamese language.
Given our focus on Vietnamese language processing, we employ PhoBERT as the pre-trained foundation for the sequential IOB tagging problem, which is subsequently adapted to accommodate streaming requirements.
\subsection{Weighted Finite-State Transducers}
Traditional rule-based approaches to ITN often struggle with the complexity and variability of natural language. To address these limitations, Weighted Finite-State Transducers (WFST) \cite{Mohri2004} have emerged as a powerful and flexible framework. WFST enables the representation of complex linguistic transformations as a network of states and transitions, where weights can encode probabilities or costs associated with different normalization options. This allows for efficient and robust ITN, capable of handling ambiguities and variations in spoken language.
 
\subsection{Model Architecture}
 \begin{figure}[ht]
\vspace{-15pt}
  \centering
  \includegraphics[width=\linewidth]{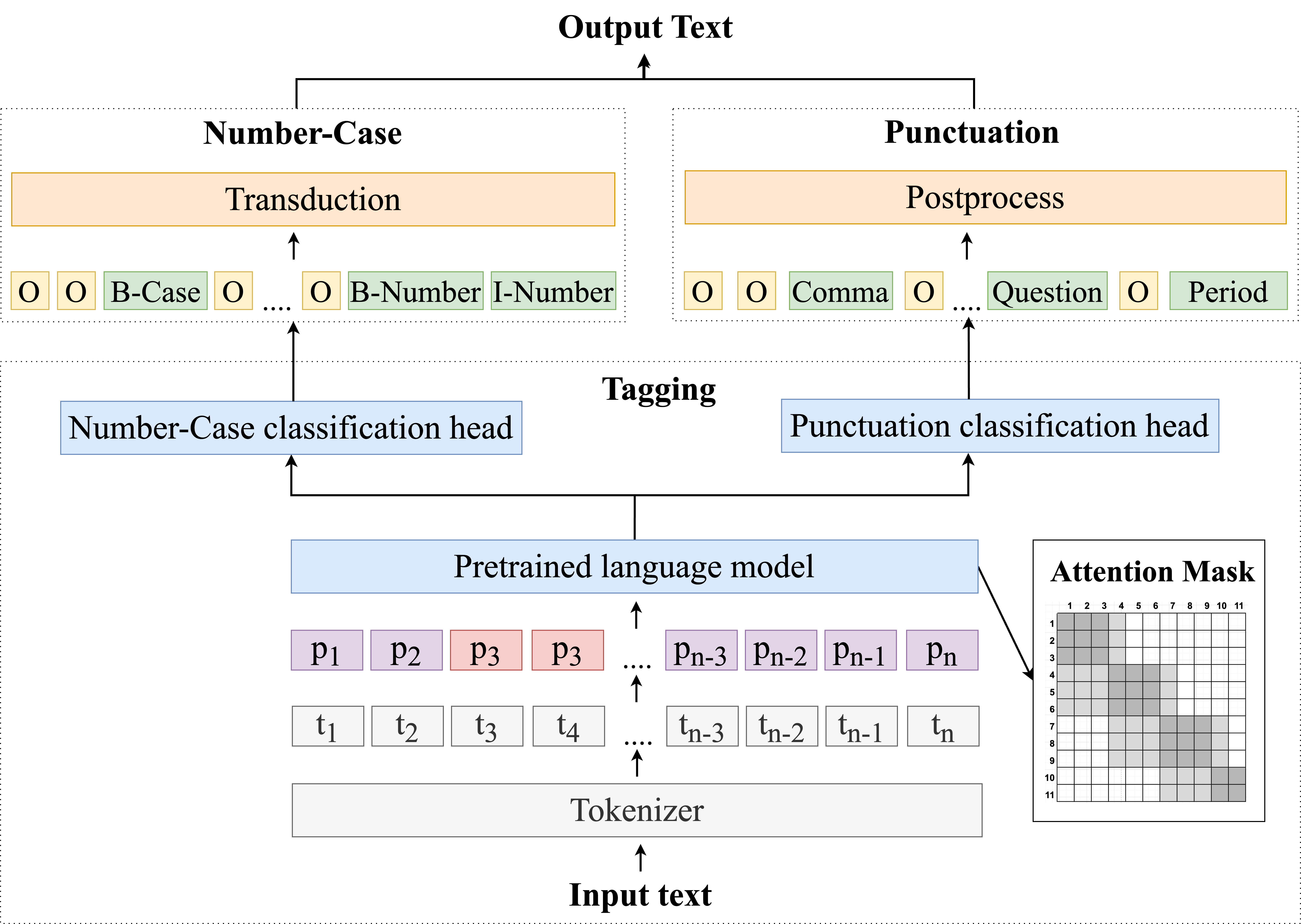}
\vspace{-15pt}
  \caption{Streaming Pretrained Language Model for Inverse Text Normalization.}
\label{fig:model}
\vspace{-15pt}
\end{figure}

The proposed streaming ITN model, as illustrated in Figure \ref{fig:model}, comprises two distinct phases: Tagging and Transduction/Postprocess. In the Tagging phase, a pre-trained BERT model serves as a contextual encoder, extracting relevant linguistic information. We employed multitask training with two distinct classification heads: one for Number-Case, categorizing tokens into four classes (Case, Number, Date, and Phone), and another for Punctuation, classifying tokens into four types (Comma, Exclamation, Period, and Question). Each classification head comprised a dropout layer followed by a dense layer, assigning each input token to one of its respective four categories. In the subsequent Transduction phase, the tagged spans produced by the Number-Case classification task are processed using WFST grammar. This grammar transforms the identified spans into their canonical written forms, facilitating real-time conversion from spoken to written text. Additionally, the output of the Punctuation classification task is handled in the Postprocess phase, where appropriate punctuation marks are inserted into the output sequence based on the predicted tags.

A considerable challenge arises from the tokenization inherent in pre-trained language models. To maintain effective word-level inference despite subword tokenization, careful consideration of subword structure is essential since a single word ``welcome'' may be tokenized as two distinct tokens, ``wel'' and ``come''. To address this, we implement a specific training strategy for label assignment. Only the first token index of a word receives the assigned label, while subsequent token indexes are masked with negative infinity values to prevent their influence on the model's output.

Finally, the model is optimized via the following loss:
\begin{equation}
\mathcal{L}_{\text{total}} = \mathcal{L}_{\text{num-case}} + \mathcal{L}_{\text{punct}}
\end{equation}
\begin{equation}
\mathcal{L}_{\text{num-case}} = -\frac{1}{N_{nc} \cdot T_{nc}} \sum_{n=1}^{N_{nc}} \sum_{t=1}^{T_{nc}} \sum_{c=1}^{C} y_{n,t,c}^{nc} \log(\hat{y}_{n,t,c}^{nc})
\end{equation}
\begin{equation}
\mathcal{L}_{\text{punct}} = -\frac{1}{N_p \cdot T_p} \sum_{n=1}^{N_p} \sum_{t=1}^{T_p} \sum_{p=1}^{P} y_{n,t,p}^{\text{p}} \log(\hat{y}_{n,t,p}^{\text{p}})
\end{equation}
Here, $N$ represents the batch size, $T$ the time steps, and $C=P=4$ the number of classes for both Number-Case ($nc$) and Punctuation categories, with $y$ denoting true labels and $\hat{y}$ predicted probabilities.

\subsection{Positional Encoding For Multi-token Words}
To efficiently capture the positional context of multi-token words generated by subword tokenization, we maintain a word-level ordering. In this scheme, words decomposed into multiple subword tokens are assigned a single, unique positional encoding, as shown in the red boxes of Figure \ref{fig:model}. Thus, the subwords `wel'' and ``come'', representing ``welcome'', share a common positional encoding. This ensures that the positional encoding reflects the word's position rather than the individual subword positions. 

\subsection{Dynamic Context-Aware Multihead Attention}

\begin{table*}[ht]
\vspace{-15pt}
\caption{Evaluation of ITN performance under various setting configurations. Model types include Non-Stream ($NS$) and Stream ($S$). Parameters $l$, $c$, and $r$ represent left context, chunk, and right context sizes at inference, respectively.}
\vspace{-4pt}
\label{tab:my_table1}

\centering
\begin{tabular}{l|c|c|c|c|c|c|c|c|c|c}
\toprule
\textbf{Model} & \textbf{Pretrained} & \multicolumn{1}{c|}{\textbf{Mask}}  & \textbf{\textit{l}} & \textbf{\textit{c}} & \textbf{\textit{r}} & \textbf{Precision} & \textbf{Recall} & \textbf{$\text{F}_1$} & \textbf{I\_WER} & \textbf{NI\_WER} \\
\midrule
$NS_1$ & $\times$ & $\times$ & $\infty$ & $\infty$ & $\infty$ & 0.76 & 0.69 & 0.72 & 7.51 & 1.72 \\
$NS_2$ & PhoBert & $\times$ & $\infty$ & $\infty$ & $\infty$ & 0.86 & 0.85 & 0.86 & 6.30 & 1.63 \\
\hline
$S_1$ \cite{gaur2023streaming} & $\times$ & Chunk Mask & 16 & \{3, 4, 5\} & 0 & 0.64 & 0.57 & 0.59 & 13.74 & 2.54 \\
$S_2$ & PhoBert & Chunk Mask & 16 &  \{3, 4, 5\}  & 0 & 0.69 & 0.68 & 0.68 & 11.99 & 2.43 \\
$S_3$ & PhoBert & Chunk Mask& 16 & \{3, 4, 5\} & \{1, 2\} & 0.74 & 0.73 & 0.73 & 8.66 & 2.22 \\
$S_4$ & PhoBert & Dynamic Context Mask & 16 & \{3, 4, 5\} & \{1, 2\} & 0.81 & 0.76 & 0.78 & 8.24 & 1.79 \\
\bottomrule
\end{tabular}
\vspace{-13pt}
\end{table*}
In speech recognition modeling, the benefits of expanding context, particularly by incorporating future information, have been thoroughly investigated in \cite{le24_interspeech, 10888640}. For streaming text modeling, we adopt the Dynamic Right Context masking technique introduced in \cite{le24_interspeech} during training. This method incorporates variable chunk and right context sizes into the attention mask, simulating the availability of right context during inference. However, unlike speech recognition tasks addressed in \cite{le24_interspeech}, our ITN system utilizes significantly smaller chunks and right context sizes. Moreover, we leverage the flexibility to employ variable chunk sizes during inference, adapting to the number of output words from each speech chunk. This adaptability overcomes the limitations of fixed chunk approaches in  \cite{gaur2023streaming} and leads to improved performance. Importantly, our tagging process operates at the word level. Therefore, when constructing the attention mask for multi-token words, we ensure the attention span aligns with the word order, mirroring the positional encoding layer.

One of the primary difficulties in integrating streaming ITN into streaming ASR systems is the problem of incomplete words at chunk boundaries. Due to disparities between ITN and ASR tokenizers, the output in an ASR chunk may be partially decoded. This fragmentation can lead to incorrect ITN inputs and outputs. For instance, if the intended phrase is ``vincom ocean park'', the first ASR chunk might yield ``Vincom o'', and the second ``cean park'', resulting in the erroneous ITN output ``Vincom oCean park'' due to incomplete word input. At inference, a right context buffer is utilized to facilitate revision. ASR outputs, including incomplete final words, are processed immediately by the ITN. However, the final word from the previous ASR chunk is retained in the buffer and appended to the current chunk's decoded results, enabling re-correction with the additional context. Consider the first chunk, ``vincom o''. The ITN processes it, and ``o'' is placed in the buffer. When the next chunk, ``cean park'', arrives, ``o'' is concatenated, allowing the ITN to recognize ``ocean'' as a complete word and generate the accurate output ``Ocean park''. 

In conclusion, our streaming ITN model benefits from its ability to manage variable context sizes and incorporates contextual awareness buffering for revision.
\section{Experiments}
\subsection{Dataset}
We employ two proprietary Vietnamese datasets, Number-Case and Punctuation, for multitask learning. Each dataset, composed of 50,000 sentences collected from a variety of sources, is partitioned into 40,000 samples for training, 5,000 for validation, and 5,000 for testing. A team of five Vietnamese native labelers manually annotates each sample, following established guidelines to maintain annotation consistency. In the Number-Case task, each word within a sentence is assigned a tag from the set {B, I, O}, where ``B'' signifies the beginning of a designated category, ``I'' indicates a word within a designated category, and ``O'' denotes words outside any designated category. For the Punctuation task, we address the placement of four common punctuation marks: commas, periods, question marks, and exclamation marks. If a word is immediately followed by one of these punctuation marks, it is labeled with that punctuation mark; otherwise, it is labeled ``O''. To enhance the trustworthiness of the dataset, each sample is independently annotated by two annotators.
\subsection{Setup}
\textbf{Training:} All models were trained for up to 200 epochs on a single NVIDIA 4090 GPU, using the AdamW optimizer. Key parameters included: learning rate of $2e-4$ with exponential decay (factor of 0.999875 per epoch), $\beta_1 = 0{.}8$, $\beta_2 = 0{.}99$, weight decay $\lambda = 0{.}01$, and a batch size of 32. Streaming ITN models utilized dynamic context masking with variable chunk sizes $\{3, 4, 5, 6, 7\}$ and right context sizes $\{1, 2\}$, while maintaining a fixed left context of 16 for both training and inference. \\
\textbf{Evaluation: } We evaluated tagging model performance using standard metrics: Precision, Recall, and F$_1$-score. 
To further assess generated text quality, especially the performance of our WFST module, we also employed ITN Word Error Rate (I-WER) and Non-ITN Word Error Rate (NI-WER), as introduced in \cite{sunkara2021neural}. I-WER measures the quality of ITN, while NI-WER quantifies unintended perturbations to the original source text.
For streaming ITN benchmarking, we simulated the streaming process using variable chunk sizes, randomly selected from 3 to 5 words, and right context sizes, randomly selected from 1 to 2 words. To guarantee reliability, the final results were averaged over five inference runs.
To evaluate the streaming capabilities of our streaming ITN model, we assessed its latency and real-time factor (RTF) on an NVIDIA GeForce RTX 4090 with 24 cores CPU Intel. For a seamless integration of ITN within a streaming ASR pipeline, the combined system must ensure that the processing latency does not exceed the duration of the input speech chunk. Consequently, both the ASR and ITN components should exhibit an RTF significantly lower than 1, guaranteeing real-time performance. We employed a Conformer model for ASR, configured with 12 encoder layers, each with 448 hidden units and 4 self-attention heads. The feed-forward layer within each encoder block has an output dimension of 1792. \\
\textbf{Statistical evaluation:} We employed a bootstrap method with 10000 iterations to determine 95\% confidence intervals for the metric of interest, leveraging the \cite{Confidence_Intervals} toolkit. Results are presented as $\textit{mean}_{[\textit{min}, \textit{max}]}$, where $\textit{mean}$ is the metric's value on the original test set, and $[\textit{min}, \textit{max}]$ represents the confidence interval's lower and upper bounds.
\subsection{Results}
\begin{table}[t]
\vspace{-18pt}
\centering
\caption{Example utterances from ITN outputs}
\label{tab:neural_itn_examples}
\vspace{-3pt}
\begin{tabularx}{0.48\textwidth}{>{\RaggedRight\arraybackslash}X}
\toprule
\textbf{Reference:} Mai ơi tiền gạo 3.2 triệu nhá \\
\textbf{Input:} mai ơi tiền gạo ba triệu hai nhá \\
$S_2$: \textcolor{red}{mai} ơi tiền gạo \textcolor[rgb]{0,0.5,0}{3.2} triệu nhá \\
$S_4$: \textcolor[rgb]{0,0.5,0}{Mai} ơi\textcolor[rgb]{0,0.5,0}{[,]} tiền gạo \textcolor[rgb]{0,0.5,0}{3.2} triệu nhá \\
\midrule
\textbf{Reference:} số 0319678116 không biết là đã thay đổi chưa? \\
\textbf{Input:} số không ba ba chín sáu bảy tám một một sáu không biết là đã thay đổi chưa \\
$S_2$: số \textcolor{red}{03196781160} biết là đã thay đổi chưa \\
$S_4$: số \textcolor[rgb]{0,0.5,0}{0319678116} không biết là đã thay đổi chưa\textcolor[rgb]{0,0.5,0}{[?]} \\
\midrule
\textbf{Reference:} ông Trung chuyển 5.8 triệu \\
\textbf{Input:} ông trung chuyển năm triệu tám \\
$S_4$: ông \textcolor{red}{trung} chuyển \textcolor[rgb]{0,0.5,0}{5.8} triệu \\
$NS_2$: ông \textcolor[rgb]{0,0.5,0}{Trung} chuyển \textcolor[rgb]{0,0.5,0}{5.8} triệu \\
\midrule
\textbf{Reference:} mai về đám giỗ dượng Mười nha, dì Mười kêu \\
\textbf{Input:} mai về đám giỗ dượng mười nha dì mười kêu \\
$S_4$: mai về đám giỗ dượng \textcolor[rgb]{0,0.5,0}{Mười} nha dì \textcolor[rgb]{0,0.5,0}{mười} kêu \\
$NS_2$: mai về đám giỗ dượng \textcolor[rgb]{0,0.5,0}{Mười} nha\textcolor[rgb]{0,0.5,0}{[,]} dì \textcolor[rgb]{0,0.5,0}{Mười} kêu \\
\bottomrule
\end{tabularx}
\vspace{-16pt}
\end{table}
A comprehensive evaluation of our proposed methods across various assessment metrics is presented in Table \ref{tab:my_table1}. Notably, we observed a significant performance enhancement when employing a pretrained model compared to a non-pretrained model, in both non-streaming ($NS$) and streaming ($S$) scenarios. 
Specifically, the  $\text{F}_1$-score of $NS_2$ and $S_2$ surpassed that of $NS_1$ and $S_1$ by \(14_{[13.1, 14.1]}\%\) and \(9_{[8.3, 9.4]}\%\) under identical settings, respectively, consistently improving both Precision and Recall. This improvement emphasizes the benefits of pretrained models for sequence tagging, particularly in tasks where contextual understanding is vital. The enriched linguistic representations derived from large-scale pretraining contribute to improved tagging accuracy and, consequently, more reliable ITN even in a real-time scenario.

However, switching from full context ($NS_2$) to limited context ($S_2$) for streaming ability resulted in a substantial performance drop, with up to a 18\% decrease in F$_1$-score. This indicates that the lack of context significantly negatively impacts ITN, even with a powerful Pretrained Language Model. 
Integrating the right context ($S_3$) resulted in a performance improvement of up to \(5_{[4.9, 5.7]}\%\) compared to ($S_2$). Notably, applying dynamic context masking ($S_4$) during training further enhanced performance by \(5_{[4.7, 5.6]}\%\), as it closely simulated the inference process. As a result, an F$_1$-score of 78\% was attained, underscoring the value of aligning training and inference conditions. These findings highlight the critical importance of right context in both training and inference. The additional contextual information provided by right context facilitates more accurate tagging decisions, minimizing ambiguities and bolstering overall robustness. Moreover, the significant performance gains achieved with dynamic context masking demonstrate its effectiveness in enabling the model to adapt to varying context lengths, thereby improving generalization in many cases. Furthermore, the 95\% confidence intervals consistently demonstrate that the observed performance changes are statistically significant and reliable. The narrow ranges around the reported improvements and drops indicate a high degree of certainty that these effects are not due to random chance. This reinforces the conclusions about the benefits of pretraining, the negative impact of limited context, and the positive contributions of right context and dynamic masking.
The results on I-WER and NI-WER further demonstrate the effectiveness of our methods and WFST system, particularly in streaming scenarios. Our optimized streaming model ($S_4$) achieves a NI-WER of 1.79, comparable to the best non-streaming model ($NS_2$), while maintaining a competitive I-WER of 8.24. Notably, ($S_4$) outperforms other streaming models that do not employ pretraining, right context inclusion, or dynamic context masking. This performance highlights its ability to accurately normalize text in a streaming fashion while preserving the integrity of the original source text.

Table \ref{tab:my_table2} presents the F$_1$ scores for our ITN models across Number-Case categories, demonstrating the impact of pre-trained language models and dynamic context awareness.  Notably, the Phone category exhibits a significant performance decrease when transitioning from full context ($NS_2$) to limited context (($S_1$) and ($S_2$)). This indicates that accurate phone number prediction heavily relies on contextual information, especially the right context, where subsequent digits are crucial for boundary and structure identification.
\begin{table}[t]
\vspace{-18pt}
\caption{$\text{F}_1$ Scores of ITN models across Number-Case classes.}
\vspace{-3pt}
\label{tab:my_table2}
\centering
\begin{tabular}{lccccc}
\toprule
\textbf{Model} & \textbf{Case} & \textbf{Date} & \textbf{Number} & \textbf{Phone} \\
\midrule
$NS_2$ & 0.82 & 0.89 & 0.91 & 0.89 \\
$S_1$  & 0.68 & 0.72 & 0.77 & 0.48 \\
$S_2$ & 0.72 & 0.76 & 0.80 & 0.56 \\
$S_4$ & 0.80 & 0.86 & 0.89 & 0.70 \\

\bottomrule
\end{tabular}
\end{table}
Similarly, Table \ref{tab:my_table3} details the evaluation of ITN models on specific punctuation types within the Punctuation category. This analysis highlights the challenges faced by streaming ITN models when predicting punctuation in limited context scenarios, with significant drops for periods, exclamation, and question marks. This disparity likely stems from the fact that these latter punctuation types often depend on discourse-level information or sentence structure that extends beyond the immediate preceding context, further emphasizing the limitations of limited context processing for certain punctuation tasks. Illustrative examples of ITN output utterances are provided in Table \ref{tab:neural_itn_examples}.
\begin{table}[t]
\vspace{-2pt}
\caption{$\text{F}_1$ scores of ITN models across Punctuation types.}
\vspace{-4pt}
\label{tab:my_table3}
\centering
\begin{tabular}{lccccc}
\toprule
\textbf{Model} & \textbf{Comma} & \textbf{Exclamation} & \textbf{Period} & \textbf{Question} \\
\midrule
$NS_2$ & 0.83 & 0.80 & 0.76 & 0.68  \\
$S_1$  & 0.48 & 0.59 & 0.33 & 0.19  \\
$S_2$ & 0.62 & 0.69 & 0.45 & 0.36  \\
$S_4$ & 0.75 & 0.46 & 0.56 & 0.58  \\
\bottomrule
\end{tabular}
\vspace{-13pt}
\end{table}
\subsection{Latency}
\begin{table}[h]
\vspace{-8pt}
\caption{Latency and RTF of streaming ITN model with ASR input chunk of 400ms and ITN right context of 1. To simulate the concurrent users (CCU), we use a batch size of 100 for both the ASR and ITN models.}
\vspace{-3pt}
\label{tab:my_table4}
\centering
\begin{tabular}{cccc}
\toprule
\textbf{Chunk size} & \textbf{ITN Latency} & \textbf{ASR Latency} & \textbf{RTF} \\
\midrule
3 & 6.53 &  & 0.137 \\
4 & 6.85 & 48.32 & 0.138 \\
5 & 6.93 & & 0.138 \\
\bottomrule
\end{tabular}
\vspace{-5pt}
\end{table}
As evidenced by Table \ref{tab:my_table4}, our streaming ITN achieves exceptional latency performance, with a maximum of 6.85 ms for 5-word chunks. The resulting RTF, significantly less than 1, validates the model's real-time processing capabilities and its ability to efficiently serve 100 concurrent users in large-scale deployments.

\section{Conclusion}
In this paper, we introduced a robust and efficient streaming Inverse Text Normalization system that effectively leverages a pretrained language model to significantly enhance tagging accuracy. By adapting a pretrained model for streaming applications, we addressed the inherent limitations of traditional ITN methods, particularly in real-time scenarios. Our key contribution lies in the development of a dynamic context-aware training and inference strategy. This approach, which incorporates dynamic chunk masking and right-context information, allows the model to adapt seamlessly to variable chunk sizes and significantly improving performance in limited-context scenarios while maintaining exceptionally low latency, crucial for real-time streaming ASR and a seamless user experience.

\nocite{*}
\bibliographystyle{IEEEtran}
\bibliography{mybib}

\end{document}